\documentclass[preprint,12pt]{elsarticle}

\usepackage{amssymb}
\usepackage{multirow}

\journal{Nuclear Physics B}

\begin{document}

\begin{frontmatter}



\title{TinySiamese Network for Biometric Analysis}


\author[label1]{Islem Jarraya\corref{mycorrespondingauthor}}

\affiliation[label1]{organization={REGIM-Lab.: REsearch Groups in Intelligent Machines, University of Sfax, National Engineering School of Sfax (ENIS), BP 1173},
            city={Sfax},
            postcode={3038}, 
            country={Tunisia}}

\author[label1,label2]{Tarek M. Hamdani}
\affiliation[label2]{organization={Higher Institute of Computer Science Mahdia (ISIMa), University of Monastir},country={Tunisia}}
\author[label3]{Habib Chabchoub}
\affiliation[label3]{organization={College of Business, Al Ain University of Science and Technology, Abu Dhabi},country={United Arab Emirates}}
\author[label1,label4]{Adel M. Alimi}
\affiliation[label4]{organization={Epartment of Electrical and Electronic Engineering Science, Faculty of Engineering and the Built Environment, University of Johannesburg},country={South Africa}}
\cortext[mycorrespondingauthor]{Corresponding author}

\begin{abstract}
Biometric recognition is the process of verifying or classifying human characteristics in images or videos. It is a complex task that requires machine learning algorithms, including convolutional neural networks (CNNs) and Siamese networks. Besides, there are several limitations to consider when using these algorithms for image verification and classification tasks. In fact, training may be computationally intensive, requiring specialized hardware and significant computational resources to train and deploy. Moreover, it necessitates a large amount of labeled data, which can be time-consuming and costly to obtain. The main advantage of the proposed TinySiamese compared to the standard Siamese is that it does not require the whole CNN for training. In fact, using a pre-trained CNN as a feature extractor and the TinySiamese to learn the extracted features gave almost the same performance and efficiency as the standard Siamese for biometric verification. In this way, the TinySiamese solves the problems of memory and computational time with a small number of layers which did not exceed 7.
It can be run under low-power machines which possess a normal GPU and cannot allocate a large RAM space. Using TinySiamese with only 8 GO of memory, the matching time decreased by 76.78\% on the B2F (Biometric images of Fingerprints and Faces), FVC2000, FVC2002 and FVC2004 while the training time for 10 epochs went down by approximately 93.14\% on the B2F, FVC2002, THDD-part1 and CASIA-B datasets. The accuracy of the fingerprint, gait (NM-angle 180°) and face verification tasks was better than the accuracy of a standard Siamese by 0.87\%, 20.24\% and 3.85\% respectively. TinySiamese achieved comparable accuracy with related works for the fingerprint and gait classification tasks.
\end{abstract}




\end{frontmatter}


\section{Introduction}
Deep learning has been successfully used to achieve good performances in a variety of applications such as recognition applications including image, activity and voice recognition. However, most of these algorithms often perform well when the predictions are made using a bit of information or data. In fact, in the modern deep learning tasks, neural networks are almost effective, but these networks require training on a huge number of samples. Nevertheless, a large number of samples is not available in some problems. For certain tasks including fingerprint or signature verification, data is not abundantly available.\\
The lack of data can be solved using such procedures as data augmentation which has drawbacks, among which changing the right direction of learning or falling into too much data. As a result, systems that incorporate these procedures tend to excel in similar cases, but sometimes fail to offer robust solutions.\\
A particularly interesting task when there are a few examples for training from scratch of each class before making a prediction is verification. This is called \textit{one-shot learning} \cite{koch2015siamese}, which attempts to solve such problems and construct a trained model using a small number of samples. One-shot algorithms can use few training samples for each class and generalize a model for unfamiliar categories without the need for extensive retraining. Therefore, the objective is to achieve new samples for each epoch and get high recognition accuracy with limited data.\\
For one-shot learning, approaches such as deep Siamese networks \cite{bromley1993signature} were proposed in several works \cite{Attrich2021,zhu2021,li2022}. Deep Siamese networks solve this type of tasks using only few images to get better predictions. The capacity to learn from little data has made Siamese networks popular in recent years. These networks are composed of two twin networks for image similarity computation. In fact, the Siamese network integrates CNN neural networks such as ResNet, VGG and MobileNet. It tries to learn the similarity between given images using distance measures.\\
Despite the effectiveness of the Siamese Networks, they possess some weaknesses. Since they require pairs to learn, these networks need more training memory and time than normal CNN networks. In fact, they are slower than the normal learning types. In addition, the Siamese networks always require a lot of running time for the prediction. 

This work aims to address these high-level problems without requiring expensive learning which may be impossible due to limited data, low-power machines or the demand for fast prediction in terms of execution time.\\
The present work work relies on the deep learning framework, which uses a small number of layers 
with relevant features to avoid learning failure. The proposed model is able to learn from little data although the cost of the learning algorithm in terms of execution time, running memory, and number of layers cannot be high. In fact, this paper introduces the TinySiamese network which has a small number of layers 
and allows a small model to successfully learn from few examples with a short learning time. TinySiamese is very useful under low-power machines with a normal GPU which cannot allocate a large RAM space.\\ 
The most important contributions of our work are the following:
\begin{itemize}
\itemsep=-1pt
\itemindent=-3pt 
\item Proposing the TinySiamese for verification using a few number of layers and without the need for a huge dataset for training.
\item Proving  that  the  proposed Tiny-Siamese could get a competitive performance with the standard Siamese Network. In the experiments, the efficiency of TinySiamese was demonstrated for verification and classification with the shorted matching and training time.
\item Showing through experimental studies that TinySiamese achieved a competitive performance with related works for classification.
\end{itemize}
Here is how the remaining part of the paper is organized: related works are presented in section 2. However, the used Siamese and the proposed TinySiamese are described in Section 3 and Section 4 respectively. Section 5 is devoted to the presentation of used datasets, whereas Section 6 focuses on the experimental study. Section 7 covers the discussion. The conclusion of this paper is drawn in Section 8.
\section{Related Works}
Research into one-shot learning algorithms is somewhat mature. It has received attention from the machine learning community. In fact, there are different works that present the Siamese as a network for one-shot learning. This section briefly reviews the related works on Siamese-based biometric recognition.

The Siamese neural network is not very different from the traditional Convolutional Neural Networks. It takes images as input and encodes their features into a set of layers. The difference lies in the output processing. In fact, the Siamese networks take 2D images as input. The similarity between the two inputs is calculated using a distance between their feature vectors. Different distances formula were used on several works as Euclidean and contrastive losses \cite{koch2015siamese}. 
The standard Siamese was performed in different works which focus on fingerprint recognition such as Lin \textit{et al.} \cite{lin2019}, Attrich {et al.} \cite{Attrich2021}, Zhu {et al.} \cite{zhu2021}, Alrashidi \textit{et al.} \cite{Alrashidi2021}, Li {et al.} \cite{li2022}, Zihao \textit{et al.} \cite{Zihao2022} and Zhengfang \textit{et al.} \cite{zhengfang2022} and on face recognition, including Song \textit{et al.} \cite{face1},  Soleymani \textit{et al.} \cite{face2}, Pei \textit{et al.} \cite{face3}, Li \textit{et al.}\cite{Li2022CatFR} and Wang \textit{et al.} \cite{face2023}. Siamese networks have also been used in other studies related to gait recognition, among which Zhang \textit{et al.} \cite{zhang2016siamese}, George \textit{et al.} \cite{george2020robust}and Sheng \textit{et al.} \cite{SHENG2020}.

Lin \textit{et al.} \cite{lin2019} developed a framework of multi-Siamese using fingerprint minutiae, respective ridge map and specific region of ridge map. Each single CNN of the Siamese network was composed of four convolutional layers and three max pooling layers. This network was used to calculate similarity scores of two fingerprint images using a distance-aware loss function.
Attrich {et al.} \cite{Attrich2021} proposed a Siamese network to develop a contactless fingerprint recognition system. The backbone of the proposed network was composed of three layers of convolutional and batch normalization and one average pooling layer. Attrich {et al.} employed the contrastive loss for the calculation of the similarity score.
Zhu {et al.} \cite{zhu2021} presented three different Siamese networks for fingerprint recognition. The three Siamese networks were based on three CNN networks: their own CNN, AlexNet and VGG. 
Li {et al.} \cite{li2022} created an embedded image processing algorithm based on a Siamese neural network for fingerprint image recognition from any source. The Siamese backbone was constructed of thirteen convolutional layers. The contrastive loss function was used in the distance layer. Alrashidi \textit{et al.} \cite{Alrashidi2021} proposed to use a Siamese network for cross-sensor fingerprint matching and enhancement. The proposed Siamese network was composed by four Convolutional (Conv) and Rectified linear unit activation (ReLU) layers among which there are three Softmax layers. Alrashidi \textit{et al.} computed the channel-wise difference between the feature maps F1 and F2 of two fingerprint images in order to determine whether or not the corresponding features in each fingerprint are similar. Zihao \textit{et al.} \cite{Zihao2022} proposed a novel fingerprint recognition method based on a Siamese neural network which was made up of thirteen convolutional layers for each sub-network. They used the contrastive loss function for the distance layer of the Siamese network. Zhengfang \textit{et al.} \cite{zhengfang2022} presented a novel Siamese Rectangular Convolutional Neural Network (SRCNN) for fingerprint identification. Each sub-network of the SRCNN is composed of five convolutional and pooling layers and each pooling layer is positioned after a convolutional layer. At the end of the SRCNN, the euclidean distance was used for the distance layer.\\
Song \textit{et al.} \cite{face1} designed a Siamese Neural Network based on Local Binary Pattern and Frequency Feature Perception for face recognition. They deployed the VGG16 network structure for feature extraction and the squared euclidean distance in the Siamese network. Pei \textit{et al.} \cite{face3} used a Siamese network for face spoofing detection. The proposed deep Siamese network was trained by joint Bayesian with contrastive and softmax losses.\\
George \textit{et al.} \cite{george2020robust} presented a deep Siamese convolutional neural network for gait recognition across viewpoints. The Siamese network contained three blocks which were each composed of 1 convolutional layer, 1 batch Normalization, 1 ReLU layer and 1 max pooling layer. George \textit{et al.} used the contrastive loss function to calculate the similarity between two inputs. Sheng \textit{et al.} \cite{SHENG2020} proposed a novel skeleton-based model named Siamese Siamese Denoising Autoencoder network for gait recognition. They proposed to learn a discriminative embedding vector using a Siamese architecture  based on a Deep Autoencoder (DAE) network. They employed the contrastive loss function for the distance layer of the Siamese network. Table \ref{tbl0} presents the related works which used the Siamese network.

\begin{table}[ht]
\caption{Related works that used the Siamese network.}\label{tbl0}
\centering\begin{tabular}{llllll}
\\
\hline
Ref & Year & Backbone & Distance Function & Parameter & Matching \\
 & & & & & Time\\
\hline
\cite{lin2019} & 2019 & CNN & Distance-aware loss & 22M & 0.149 ms \\
\cite{george2020robust} & 2020 & CNN & contrastive loss & - & - \\
\cite{SHENG2020} & 2020 & DAE & contrastive loss & - & -\\
\cite{Attrich2021} & 2021 & CNN & contrastive loss & 38M & - \\
\multirow{2}{*}{\cite{zhu2021}} & \multirow{2}{*}{2021} & AlexNet, & \multirow{2}{*}{-} & \multirow{2}{*}{40M} & \multirow{2}{*}{-} \\
 & & VGG-16 & & & \\
\cite{Alrashidi2021} & 2021 & CNN & Euclidean distance & - & 18,81 ms\\
\cite{Zihao2022} & 2022 & CNN & contrastive loss & - & 600 ms\\
\cite{zhengfang2022} & 2022 & CNN & Euclidean distance & - & -\\
\cite{li2022} & 2022 & CNN & contrastive loss & 37M & - \\
\cite{face1} & 2022 & VGG16 & Euclidean distance & - & -\\
\multirow{2}{*}{\cite{face3}} & \multirow{2}{*}{2022} & \multirow{2}{*}{CNN} & Bayesian + contrastive & \multirow{2}{*}{-} & \multirow{2}{*}{-} \\
 & & & + softmax losses & &  \\
\hline
\end{tabular}
\end{table}

Despite the effectiveness of the Siamese Network, it has always presented 
a long matching time. Thus, the present challenge is how to keep the efficiency of the Siamese with less 
complexity and execution time. In this context, researches in the two works \cite{tinySiam1} and \cite{tinySiam2} proposed a light siamese architectures for objects tracking with small parameters and calculations. However, inputs have to pass over the set of CNN layers of Siamese for training and matching. In fact, like all standard Siamese networks, using other descriptors or only feature vectors as inputs is hopeless.\\
The proposed TinySiamese resolved this problem and achieved the above goals with a recognition accuracy very close to the accuracy of the standard Siamese.
\section{Siamese Network}
A Siamese neural network is a category of neural network architectures. It contains two or more identical sub-networks. In fact, Siamese sub-networks have the same structure with the same parameters and weights. These networks are used to find the input similarity by calculating the distance between its feature vectors.

\begin{figure}[h]
	\centering
	\includegraphics[scale=0.45]{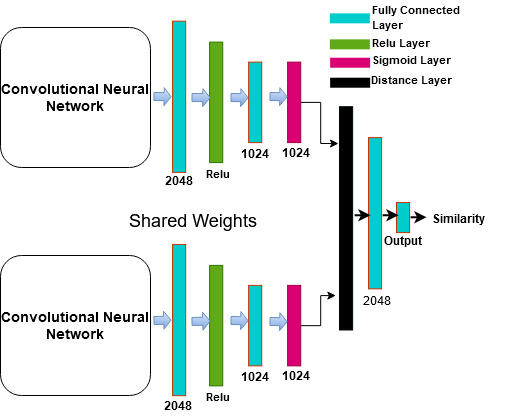}
	\caption{The Standard Siamese Network.}
	\label{FIG:Siamese2D}
\end{figure}

\begin{figure}[h]
	\centering
	\includegraphics[scale=0.45]{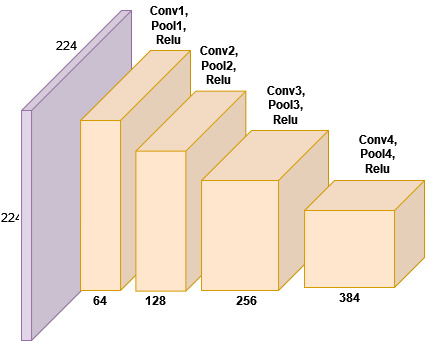}
	\caption{The Standard Siamese Network Backbone.}
	\label{FIG:cnn}
\end{figure}

In the present work, a siamese network was inspired by the well-known work of Koch \textit{et al.} \cite{koch2015siamese} and used for verification. The suggested architecture was based on a CNN sub-network, distance layer, linear layer and a sigmoid layer as shown in Fig. \ref{FIG:Siamese2D} . The CNN sub-network of the Siamese model contained four blocs of convolutional, ReLU and max Pooling layers (Fig. \ref{FIG:cnn}) ending with two fully-connected layers: ReLU and Sigmoid layers (Fig. \ref{FIG:Siamese2D}). This network was trained from scratch with the binary cross-entropy loss function and Adam optimization algorithm. The loss function is presented in the equation (Eq. \ref{Eq.BCE}) of the next section where $p_{i}$ is the predicted probability distribution, $y_{i}$ is  the actual probability distribution and N is the number of samples.
Despite the effectiveness of Siamese Network, it has some weaknesses. Since Siamese networks requires pairs to learn, it necessitates 
a large training memory and much running time for training and testing. Therefore, it requires power machines for priced learning. 
In addition, Siamese Network is limited by the CNN model using particularly image data. In other words, the freedom to use another model architecture or descriptor is generally limited.
\section{TinySiamese Network}
\begin{figure*}[h]
	\centering
	\includegraphics[scale=0.35]{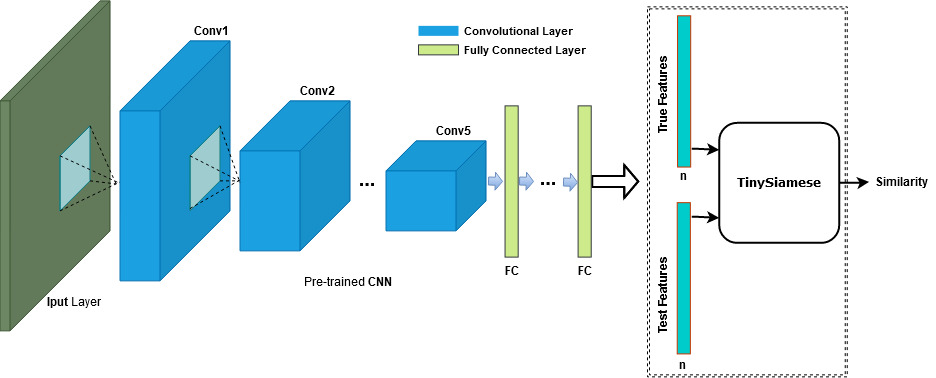}
	\caption{The Proposed Architecture Based on TinySiamese Network for Verification.}
	\label{FIG:GlobaltinySiamese}
\end{figure*}

The proposed TinySiamese neural network takes on a new look and a new way of working which is different from the standard Siamese network. The difference first appears in the input processing of the network. 
Instead of having images as input, the input was the output feature vector of a pre-trained CNN model. In other words, all input images would be transformed into feature vectors using a feature extractor (such as a pre-trained CNN model) as illustrated in Fig. \ref{FIG:GlobaltinySiamese}.
Then, the Tiny-Siamese encoded the features in a small set of layers and finally calculated the distance between two encoded feature vectors and generated similarity score. Using this score, the model was trained from scratch with the Adam optimization algorithm and binary cross-entropy loss function.
\subsection{Architecture}
\begin{figure}[t]
	\centering
	\includegraphics[scale=0.45]{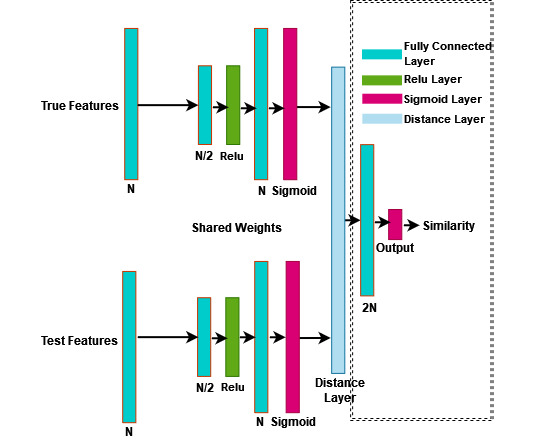}
	\caption{The proposed TinySiamese Network.}
	\label{FIG:tinySiamese}
\end{figure}

Unlike the standard Siamese, the input of the TinySiamese was the encoded image as a feature vector. The backbone layers first aimed to extract relevant features using a linear fully-connected layer and a ReLU layer and then amplify them using another linear fully-connected layer and Sigmoid layer. The output size of the first linear layer had the half size of the input (n, n/2) and was followed by a non-linear ReLU layer. The second linear layer took n/2 features in input and came back to the same first input size in output (n/2, n). This layer was followed by a non-linear Sigmoid layer. The outputs of the TinySiamese sub-networks were encoded into an n-dimensional vector using inputs of a size equal to n. Siamese networks are usually trained using a distance function to minimize the distance among matches and maximize the distance among mismatches. The Euclidean distance L2 was well performed by different works  \cite{Alrashidi2021,putra2021,face1}. Because there were no convolutional layers to lead to a strong training in the TinySiamese network, it became necessary to improve the distance function for a better separation of the classes. Thus, the Euclidean distance was merged with the Hadamard product and was used in the distance layer. The concatenated distance vector had a size equal to 2n. The final linear layer had the same size as the distance vector 2n in input and had an output of one (2n, 1). This layer was followed by a non-linear Sigmoid layer to map the produced real value into a small range that could be interpreted as probability of the similarity.\\
Fig. \ref{FIG:tinySiamese} details the overall architecture of the TinySiamese network. This model was designed to have few layers 
to be run and trained quickly even on weak machines. The proposed network was scalable depending on the size of the input feature vector and the complexity of the task. Indeed, it was possible to add new layers and delete others before and after the distance layer.
\subsubsection{Distance Layer}
Most related works used Euclidean or contrastive distances. Since the distance layer was in the middle of the network, it was more useful to use the Euclidean distance in order to receive two feature vectors (from the pair input) and output a feature vector (the difference between the two inputs). It was thought that V1 was the vector resulting from the upper sub-network and V2 was the vector resulting from the lower sub-network of the TinySiamese. In order for the distance between two feature vectors to have been more robust and efficient, the distance layer generated the concatenation of the Euclidean distance between V1 and V2 and the Hadamard product of V1 and V2. In fact, the L2 Euclidian distance and the hadamard product were used by \cite{koch2015siamese} for Authorship Verification. The resulting concatenated vector V had a 2n dimension (Eq. \ref{Eq.5}).
\begin{equation}
V=[(V1-V2)^{2} \;\ (V1 \odot V2)]
\label{Eq.5}
\end{equation}
\subsubsection{Loss Function}
The binary cross-entropy loss is a commonly used loss function in machine learning for binary classification problems \cite{fingerprint1,face5}. It measures the difference between the predicted probability distribution $p_{i}$ of the model and the actual probability distribution $y_{i}$ of the target variable as shown in the equation (Eq. \ref{Eq.BCE}) where N is the number of samples.
\begin{equation}
    Loss = \frac{1}{N}\sum_{i=1}^{N} -(y_{i} * log(p_{i}) + (1-y_{i}) * log(1-p_{i}))
    \label{Eq.BCE}
\end{equation}
\subsubsection{Activation Function}
The role of the activation function is to generate an output signal using the received input signal \cite{deep}. 
The activation function was applied to the output of any linear transformation of the proposed TinySiamese to introduce non-linearity into the model. The used activation functions were ReLU \cite{relu} and Sigmoid \cite{deep}.
The formula for the Rectified Linear Unit (ReLU) activation function is:
\begin{equation}
 f(x) = max(0,x) 
 \label{Eq.RELU}
\end{equation}
The advantage of using the ReLU activation function is that it is simple and computationally efficient. It is also the most frequently-used activation function in neural networks particularly in Convolutional Neural Networks (CNNs). Besides, ReLU can introduce sparsity in the network, which can help reduce overfitting.\\
The TinySiamese network resolves a binary classification problem by classifying two inputs into one of two classes: similar or not similar. In fact, the output result was 0 or 1. Thus, the final linear fully connected layers in the proposed TinySiamese network adopted the Sigmoid activation function (Eq.\ref{Eq.sigmoid}).
\begin{equation}
\sigma (x)=\frac{1}{1+e^{-x}}
 \label{Eq.sigmoid}
\end{equation}
\subsection{Pre-trained Model for Feature Extraction}
The convolutional neural network (CNN) is among the major architectures employed in deep learning for feature extraction. It is specially efficient type of neural network when working with visual data \cite{conv1,conv2,conv3,conv4,conv5,face3D1,facee3}. 
One of the interesting advantages of CNN is that it is able to extract features from images at various levels. A hierarchical structure of features is developed by a trained convolutional neural network with large and high-level features in the deep layers and small features in the first layers. Due to data scarcity for some tasks such as fingerprint verification \cite{fingerprint1}, it is useful to exploit the features produced by a pre-trained CNN model in different levels for TinySiamese training.\\
In other cases, when the data does not contain images (such as 3D face, video, voice, text, etc), it is possible to use any other trained model to extract features (such a swavelet neural networks, GAN, LSTM, Gabor descriptor, Autoencoder, Bert, etc) \cite{speech1,fingerprint2,gait1,gait2,gait3,facee1,facee2,face3D2,face3D3,hamdani2007multi,hamdani2011hierarchical}. In fact, the type of the feature extractor is not important knowing that the TinySiamese takes a feature vector as input. The TinySiamese is not limited by the CNN model. Furthermore, the use of any other trained model architecture or descriptor is possible.
\section{Datasets}
In the experiments, the proposed TinySiamese network was performed on four datasets:
\subsection{B2F (Biometric images of Fingerprints and Faces)}
\subsubsection{Set of Fingerprint Images}
\begin{figure}[ht]
	\centering
	\includegraphics[scale=0.2]{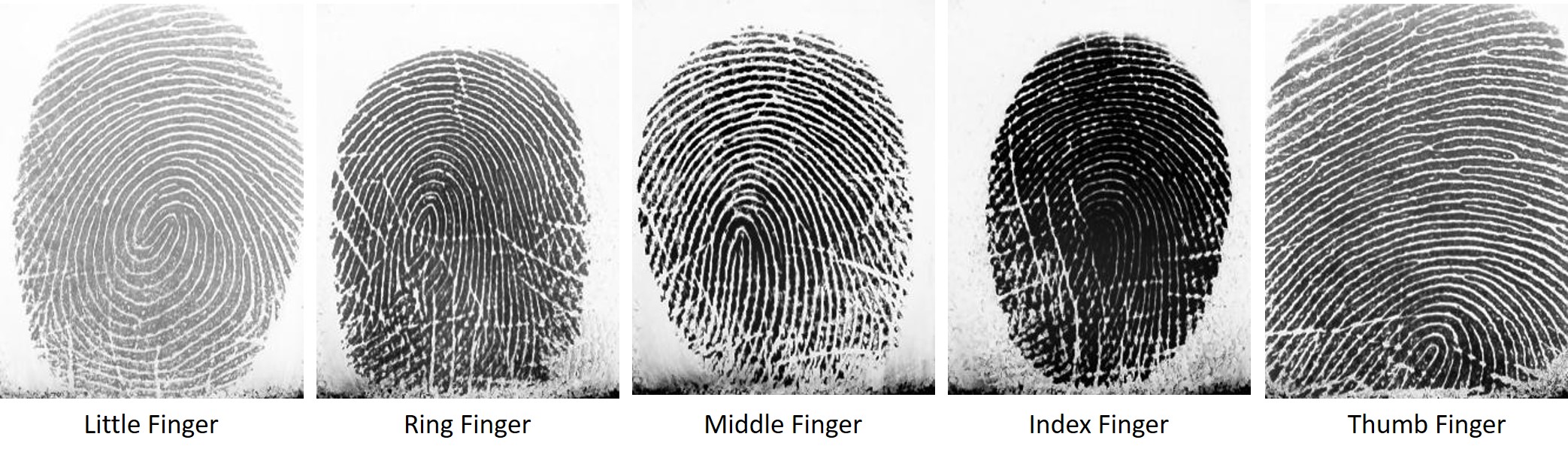}
	\caption{Fingerprint images of one hand of one person.}
	\label{FIG:fingPBMLT}
\end{figure}

To the best of our knowledge, there is no fingerprint pattern database showing the different finger image for each person. In fact, each finger is presented as a different class in most benchmark datasets. Therefore, our B2F\footnote{https://ieee-dataport.org/documents/b2f-biometric-images-fingerprints-and-faces} dataset (Biometric images of Fingerprints and Faces) has been prepared for fingerprint verification, classification or recognition. The B2F presents a multi-finger dataset. The introduced fingerprint set of B2F dataset is composed of two subsets of data: the first is for the fingerprint images of the left hand and the second is for the fingerprint images of the right hand of each person. Figure \ref{FIG:fingPBMLT} shows the fingerprint images of one hand of only one person. The digital images of this dataset were taken at a resolution of 258* 336 pixels. Each data subset (for right of left hand) contains 13.710 fingerprint images of 457 persons. In fact, there are 30 images for each hand of one person: 6 images for  each finger (little, ring, middle, index and thumb fingers).
\subsubsection{Set of Face Images}
\begin{figure}[ht]
	\centering
	\includegraphics[scale=0.3]{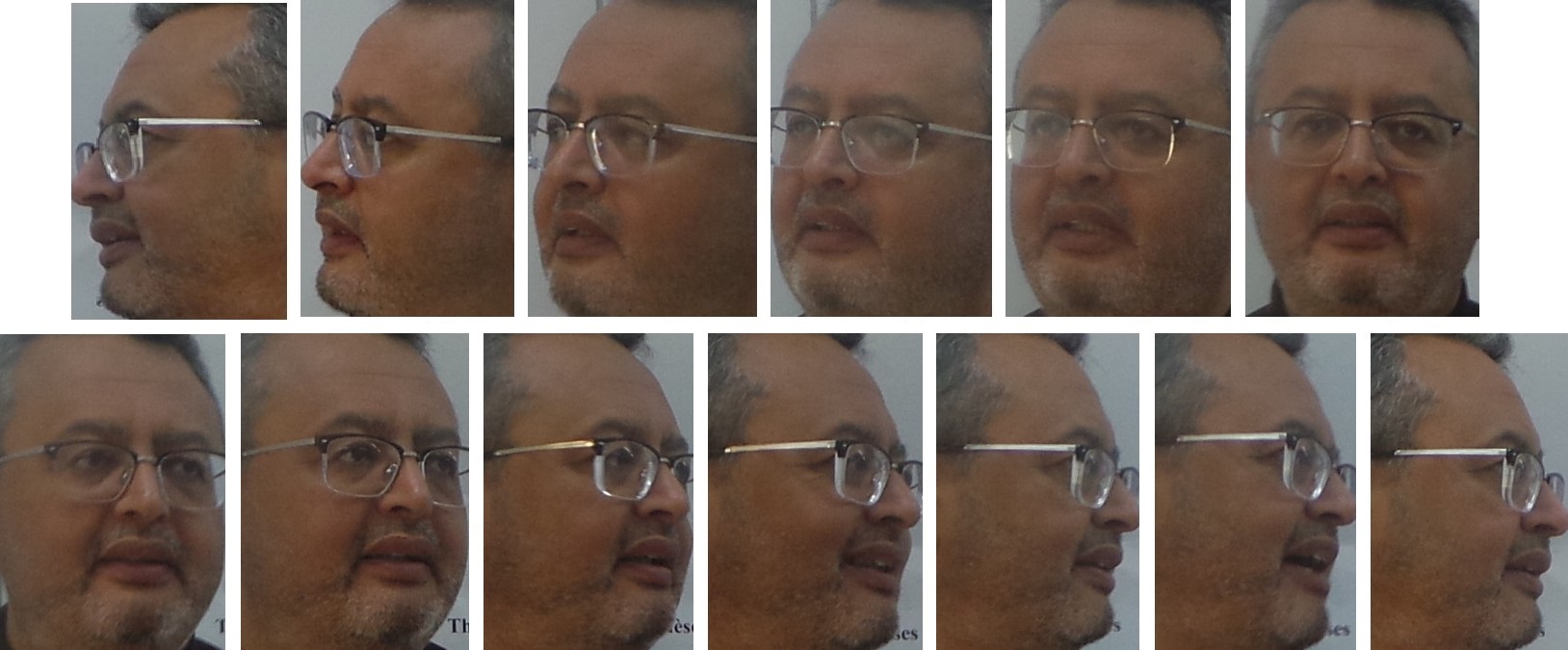}
	\caption{Face images of one person in 12 views (the last author).}
	\label{FIG:facePBMLT}
\end{figure}

The same applies to the set of face images from our B2F dataset which is composed of different facial emotions and views of the same person and distances between the camera and the person that do not exist in the published benchmarks. The introduced face set of B2F dataset is composed of face images taken on 13 views at a shooting distance of 2 meters (0°, 15°, 30°, 45°, 60°, 75°, 90°, 105°, 120°, 135°, 150°, 165° and 180°), face images taken with 7 emotions (neutral, happy, angry, disgusted, surprised, scared and sad) and four other images taken at 3 distances (2 at a distance of 2 meters, 1 at 1 meter and 1 at 0.25 meter) for each person. Figure \ref{FIG:facePBMLT} shows the face images of one person. The digital images of this dataset were taken at a resolution of 4608*3456 pixels. This dataset contains 10637 images of 431 persons. In fact, there are between 20 and 30 images for each person.
\subsection{FVC Datasets}
\begin{figure}[ht]
	\centering
	\includegraphics[scale=0.2]{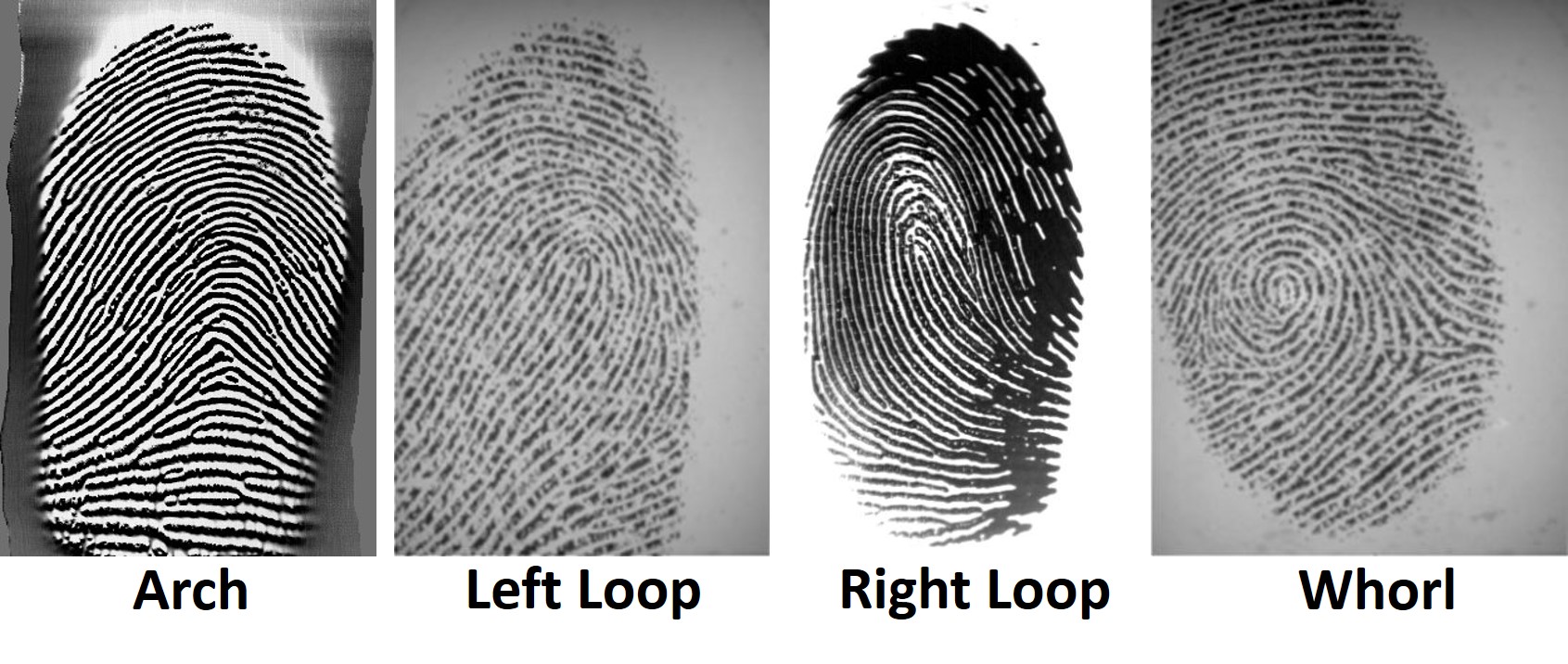}
	\caption{Fingerprint images from FVC2004.}
	\label{FIG:FingFVC}
\end{figure}

FVC2002\footnote{http://bias.csr.unibo.it/fvc2002/databases.asp} and FVC2004\footnote{http://bias.csr.unibo.it/fvc2004/download.asp} datasets include noisy images acquired by different live scan devices. Each one has four sets and contains 880 fingerprint images. The fingerprints of each dataset were categorized into four types: arch, right loop, left loop, and whorl. The four sets of FVC2004 were merged into a single set of four classes to form a multi-sensor fingerprint dataset. The same procedure was used for FVC2002 using only three sets (DB1, DB2 and DB4). Figure \ref{FIG:FingFVC} shows examples of the four classes from the FVC2004 dataset.
\subsection{THDD-part1 Dataset}
The THDD-part1 (THODBRL) \cite{jarraya2022} is a multi-view horse face database. The images were captured when horses were in the barn. The horse face images were taken from 3 views (right view profile, left view profile and frontal view of the horse). In fact, this dataset contains face images for 47 Arabian, Barbaro and hybrid horses. For each horse, there are 10 right profile images, 10 left profile images and 10 frontal face images.
\subsection{CASIA-B Dataset}
CASIA-B\footnote{http://www.cbsr.ia.ac.cn/english/Gait\%20Databases.asp} dataset is a multi-view gait database \cite{casia}. This dataset includes data of 124 persons captured from 11 views (0°, 90°, ..., 180°) with 3 walking conditions: walking with a bag (BG) (2 sequences per subject), wearing a coat (CL) (2 sequences per subject) and normal (NM) (6 sequences per subject). Namely, for each subject, there are 11 × (6 + 2 + 2) = 110 sequences. 
\begin{figure}[ht]
	\centering
	\includegraphics[scale=0.4]{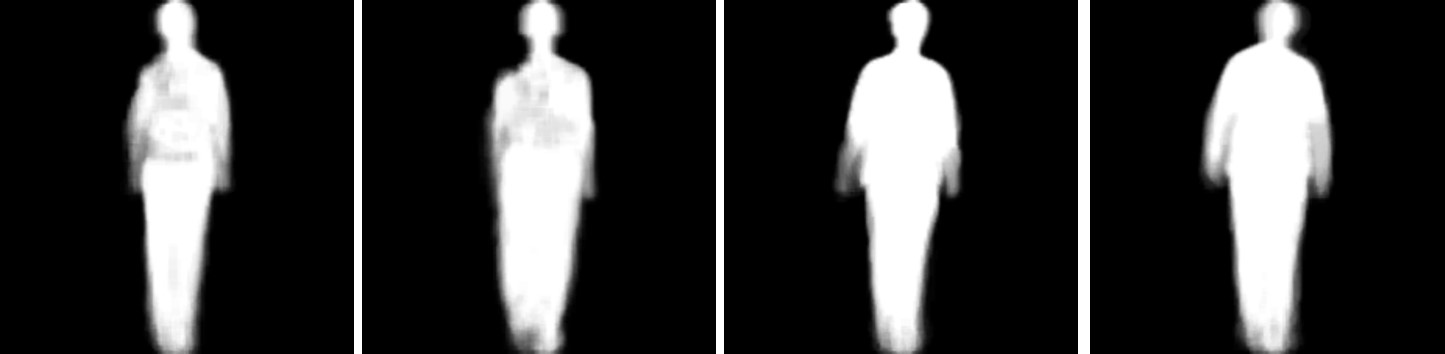}
	\caption{GEI images for four persons from CASIA-B dataset.}
	\label{FIG:gait}
\end{figure}
\section{Experiments}
\subsection{Implementation}
The experiments were performed on NVIDIA GeForce GPU and a 18 GB memory.

For the purpose of obtaining a thorough assessment, the proposed TinySiamese was tested for two different tasks on four datasets: fingerprint and face sets of the B2F , FVC , THDD-part1  and CASIA-B datasets.\\
Motivated by the fingerprint verification task (Verif), the B2F fingerprint images of the left hand dataset were divided into two different parts. The first part encompassed 85 persons. This part was intended for creating the pre-trained CNN which would serve as a feature extractor. The remaining 186 persons were for the second part. This part was randomly divided into two parts: 60\% was provided for Siamese and TinySiamese training while 40\% was devoted to testing. In fact, the fingerprint images of five fingers of each person were randomly split into two parts and used for training and testing. The TinySiamese could prove its effectiveness by implementing the introduced task which was so complex.\\
The same procedure was used for face verification, the B2F face images were merged with THDD-part1 images to get a mixed face dataset for humans and horses. The new dataset was randomly divided into two parts: 60\% was deployed for Siamese and TinySiamese training and 40\% for testing.\\
Since there was no official partition of the training and testing sets of the CASIA-B dataset, the experiments in this paper used a popular setting in current literature. This setting is also known as large-sample training (LT) when 74 subjects are used for training and the remaining 50 subjects are left for testing. The GEI was employed for gait classification. In fact, each person had four GEI images for the Gallery set and two GEI images for the Probe set. Figure \ref{FIG:gait} presents four GEI images from four videos of four different people. Only the normal state in the 180° view was used in this work for TinySiamese verification.

Motivated by the fingerprint classification task (Classif), the three datasets (DB1, DB2 and DB4) of FVC2002 were merged into one dataset and then divided into four categories for the purpose of creating the pre-trained CNN which would serve as a feature extractor. The four datasets (DB1, DB2, DB3 and DB4) of the FVC2004 dataset were merged and then categorized into four types which were, in turn, divided into two parts. According to related works, the merged dataset was randomly split: 80\% was used for training while 20\% was employed for testing. By implementing the classification task, the TinySiamese could prove its efficacy when comparing its results with the results of related works.\\
For gait classification, the same partitions deployed for verification was used again. Only the normal state in the 0°, 90° and 180° views was deplyed in our work for the evaluation of the TinySiamese classification.

The simplest CNN, AlexNet \cite{krizhevsky2017}, was employed as a feature extractor for the fingerprint verification task. AlexNet was trained using the SGD optimizer (Opt),  90 epochs (Ep) and a Batch Size (BS) equal to 5. Indeed, to make a fairer comparison between the TinySiamese and the standard Siamese network, AlexNet was fine-tuned on the second part of the B2F dataset. In order not to overuse the machine and waste more training time, 50 epochs were enough with a batch size equivalent to 5.\\
According to Ametefe et al. \cite{Ametefe2022}, the VGG16 \cite{vgg16} is effective for fingerprint classification. Thus, VGG16 was deployed as a feature extractor for the fingerprint classification task. This network was trained using the SGD optimizer (Opt), 90 epochs (Ep) and a Batch Size (BS) equal to 50.\\
The pre-trained DensNet used by Jarraya \textit{et al.} \cite{jarraya2022} was employed as a feature extractor for the gait classification task.\\
The pre-trained ResNet-50\footnote{https://awesomeopensource.com/project/peteryuX/arcface-tf2} with ArcFace \cite{deng2019arcface} was used for face verification and classification tasks. Since ResNet-50 (with ArcFace) was pre-trained only on human faces without horse faces, this model was fine-tuned on the training part of the merged face. 15 epochs with a batch size equal to 50 were enough to avoid overusing the machine and wasting more training time.

The Siamese and TinySiamese Networks were trained from scratch using ADAM optimizer, 120 epochs and a batch size equal to 18 for the fingerprint verification task. However, the same network used 240 epochs from scratch and a batch size equal to 32 for the classification task as there were more data in the FVC2002 dataset. Since each class contained hundreds of images, it was useful to increase the number of dissimilar and similar images. The same number of epochs was used for the other tasks: Gait and face verification or classification on the other datasets. 

Table \ref{tbl1} and table \ref{tbl2} represent the experimental configuration of the different networks.

\begin{table}[h]
\caption{Experimental configuration of different Networks.}\label{tbl1}
\centering\begin{tabular}{llllll}
\hline
Network & Task &Opt & BS & Ep & Loss\\
\hline
AlexNet & Fingerprint Verif & SGD & 5 & 90 & CEL \\
VGG16 & Fingerprint Classif & SGD & 5 & 90 & CEL \\
ResNet-50 + ArcFace & Face Verif/Classif & SGD & 50 & 15 & CEL\\
\hline
\end{tabular}
\end{table}

\begin{table}[h]
\caption{Experimental configuration of different Networks.}\label{tbl2}
\centering\begin{tabular}{llllll}
\hline
Network & Task &Opt & BS & Ep & Loss\\
\hline
\multirow{3}{*}{Siamese} & Fingerprint Verif & ADAM & 18 & 120 & BCEL \\
  & Gait Verif & ADAM & 6 & 120 & BCEL \\
  & Face Verif & ADAM & 16 & 120 & BCEL \\
 \hline
\multirow{4}{*}{\textbf{TinySiamese}} & Fingerprint Verif & \multirow{2}{*}{ADAM} & 18 & \multirow{2}{*}{120} & \multirow{2}{*}{BCEL} \\
 & Fingerprint Classif &  & 32 &  &  \\
  & Gait Verif/Classif & ADAM & 6 & 120 & BCEL \\
  & Face Verif & ADAM & 16 & 120 & BCEL \\
\hline
\end{tabular}
\end{table}
\subsection{Sample Pairing}
One way to use Siamese networks which consist of two identical sub-networks is to take two input samples, pass them through the sub-networks, and then compare the output representations of the samples to determine their similarity. In fact, for sample pairing, there are two different strategies. The first strategy is unbalanced sample pairing and the number of dissimilar pairs is greater than the similar pairs. The second one is the balanced pairing of samples when the number of dissimilar pairs is equal to similar pairs. The balanced pairing was used in \cite{Alrashidi2021,Li2021,Salomon2020}.

In this paper, the strategy of balanced sample pairing was followed for a stable training and testing process. For verification and classification tasks, each iteration took N similar pairs of the same user and N dissimilar pairs composed of N random user images. The same user was compared with other users for dissimilar pairs. 
\subsection{Complexity: Matching and Training Time}
The complexity of neural networks is interesting in the deep learning axis \cite{face5,jarraya2022}. Indeed, it is useful to reduce the number of layers 
in the network while maintaining performance and efficiency. Added to that, 
using a light network help users train and test deep systems, even with low-power machines. The proposed solution does not require a large Siamese network for verification. 
The standard Siamese Network can be replaced by a pre-trained CNN network and the proposed TinySiamese. The pre-trained network can be used with no or only little fine-tuning. Fine-tuning a simple CNN requires less time and memory than the simple Siamese which consists of two CNN structures in parallel. Besides, the TinySiamese 
takes a little time for matching and training. 
Table \ref{Tab:MatchTime} presents the matching time on the B2F, the FVC2000, FVC2002 and FVC2004 using Siamese network and TinySiamese. The resulting time was the average time of ten matches of ten random images. The required time for matching by the pre-trained CNN with TinySiamese is smaller than the others. In fact, there was a gain of 0.086 ms for matching.
\begin{table}[ht]
\caption{Comparison of Average Matching Time (seconds).}\label{Tab:MatchTime}
\centering\begin{tabular}{llll}
\hline
Database & Siamese & TinySiamese & Saving\\
 & & & Time (\%) \\
\hline
Casia-B (GEI) & 0.100 & \textbf{0.001} & 99.00\\
B2F (Face) & \multirow{2}{*}{0.084} & \multirow{2}{*}{\textbf{0.001}} & \multirow{2}{*}{98.81}\\
+ THDD-part1 & & &\\
B2F (Fingerprint) & 0.082 & \textbf{0.032} & 60.97\\
FVC2000 DB1 & 0.177 & \textbf{0.024} & 86.44\\
FVC2000 DB2 & 0.091 & \textbf{0.024} & 73.63\\
FVC2000 DB3 & 0.121 & \textbf{0.025} & 79.44\\
FVC2000 DB4 & 0.095 & \textbf{0.024} & 74.74\\
FVC2002 DB1 & 0.147 & \textbf{0.028} & 80.95\\
FVC2002 DB2 & 0.084 & \textbf{0.023} & 72.62\\
FVC2002 DB3 & 0.095 & \textbf{0.027} & 71.58\\
FVC2002 DB4 & 0.120 & \textbf{0.023} & 80.83\\
FVC2004 DB1 & 0.078 & \textbf{0.025} & 67.95\\
FVC2004 DB2 & 0.112 & \textbf{0.029} & 74.11\\
FVC2004 DB3 & 0.113 & \textbf{0.026} & 76.99\\
FVC2004 DB4 & 0.140 & \textbf{0.027} & 80.71\\
\hline
\end{tabular}
\end{table}
Table \ref{Tab:MatchTrain} presents the average training time of ten epochs on the fingerprint set of B2F, CASIA-B, the fusion of the face set of B2F with THDD-part1, FVC2000, FVC2002 and FVC2004 datasets using the Siamese network and the TinySiamese. The required time for training by TinySiamese was less than the standard Siamese. Indeed, there was a gain of 612.28 ms for training.
\begin{table}[ht]
\caption{Comparison of Average Training Time for 10 epochs (s).}\label{Tab:MatchTrain}
\centering\begin{tabular}{llll}
\hline
Database & Siamese & TinySiamese & Saving\\
 & ATT & ATT & Time (\%)\\
\hline
B2F (Fingerprint) & 1404.25 & \textbf{133.23} & 90.51\\
FVC2002 DB1 & 174.87 & \textbf{20.87} & 88.06\\
FVC2002 DB2 & 175.78 & \textbf{19.52} & 88.89\\
FVC2002 DB3 & 174.86 & \textbf{18.97} & 89.15\\
FVC2002 DB4 & 175.10 & \textbf{20.77} & 88.14\\
CASIA-B & 128.02 & \textbf{16.98} & 86.74\\
B2F (Face) & 2368.52 & \textbf{85.10} & 96.41\\
+ THDD-part1 & & &\\
\hline
\end{tabular}
\end{table}
\subsection{Fingerprint Verification Results}
The fingerprint verification task of five fingers at once was quite complicated. Its implementation proved the effectiveness of the proposed TinySiamese. Table \ref{Tab:verifPBMLT} shows the performance of the TinySiamese with an accuracy equal to 90.13\%. This percentage is considered good due to the difficulty of the task. In fact, by means of a pre-trained CNN, the TinySiamese was able to outperform the standard Siamese using an entire CNN for training with 0.87\%. 
\begin{table}[ht]
\caption{Verification Results on the set of Fingerprint Images of the B2F dataset.}\label{Tab:verifPBMLT}
\centering\begin{tabular}{lllll}
\hline
Method & Accuracy(\%) & Precision(\%) & Recall(\%) & F1(\%)\\
\hline
Siamese & 89.26 & 89.62 & 89.26 & 89.24 \\
\textbf{TinySiamese} & \textbf{90.13} & \textbf{91.01} & \textbf{90.13} & \textbf{90.07} \\
\hline
\end{tabular}
\end{table}
\subsection{Face Verification Results}
The face verification task using a dataset of human images on 13 views and 7 human images with 7 emotions along with horse images on 3 views was quite complicated. The implementation of this task proves again the effectiveness of the proposed TinySiamese. Table \ref{Tab:verifPBMLTFa} shows the performance of the TinySiamese with an accuracy equal to 85.87\%. This percentage is considered acceptable due to the complexity of the task. Indeed, using a pre-trained CNN, the TinySiamese was able to outperform the standard Siamese for face verification by means of an entire CNN for training with 3.85\%. 
\begin{table}[ht]
\caption{Verification Results on set of Face Images of B2F and THDD-part1 datasets.}\label{Tab:verifPBMLTFa}
\centering\begin{tabular}{lllll}
\hline
Method & Accuracy(\%) & Precision(\%) & Recall(\%) & F1(\%)\\
\hline
Siamese & 82.02  & 84.24 & 82.02 & 81.73 \\
\textbf{TinySiamese} & \textbf{85.87} & \textbf{88.36} & \textbf{85.87} & \textbf{85.64} \\
\hline
\end{tabular}
\end{table}
\subsection{Gait Verification Results}
Based on Table \ref{Tab:verifcasia180}, the proposed TinySiamese has proven its performance in the experimental part on the normal state of the CASIA-B database with the 180° view compared to the Siamese network. In fact, the  results  were  very  encouraging with an accuracy equal to 98.39\%. By means of a pre-trained CNN, the TinySiamese was able to outperform again the standard Siamese for gait verification using an entire CNN for training with an accuracy difference of 20.24\%.
\begin{table}[ht]
\caption{Verification Results on Normal set of CASIA-B dataset for the angle 180°.}\label{Tab:verifcasia180}
\centering\begin{tabular}{lllll}
\hline
Method & Accuracy(\%) & Precision(\%) & Recall(\%) & F1(\%)\\
\hline
Siamese & 78.15  & 80.62 & 78.15 & 77.70 \\
\textbf{TinySiamese} & \textbf{98.39} & \textbf{98.44} & \textbf{98.39} & \textbf{98.39} \\
\hline
\end{tabular}
\end{table}
\subsection{Fingerprint Classification Results}
The proposed TinySiamese network was tested again for a simple task on the FVC2004 benchmark. The fingerprint classification task was introduced into some research works \cite{zia2019,Nguyen2019,Saeed2022,nahar2022} using the same dataset. According to related works, the TinySiamese was trained on 80\% of the FVC2004 dataset and tested on the remaining part. The VGG16 was pre-trained using three sets of the FVC2002 (DB1, DB2 and DB4). Table \ref{Tab:compfinger} illustrates comparative results with different research studies. It is noticed that our network produced a high performance and was a competitor to the other networks.
\begin{table}[ht]
\caption{Classification results of TinySiamese on FVC2004 datasets.}
\label{Tab:compfinger}
\centering\begin{tabular}{lll}
\hline
\textbf{Ref} & \textbf{Method} & \textbf{Acc(\%)} \\
\hline
\cite{zia2019} & B-DCNNs & 95.30 \\
\hline
\cite{Nguyen2019} & CNN (3 classes) & 96.10\\
\hline
\cite{nahar2022} & LeNet (FVC2004-DB1)
 & 99.10\\
 \hline
\cite{Saeed2022} & DeepFKTNet model & 98.89\\
\hline
Our & TinySiamese & 99.00\\
\hline
\end{tabular}
\end{table}
\subsection{Gait Classification Results}
The gait classification task was introduced using the Casia-B dataset in some research works \cite{Huang2021,Huang2021(1),Lin2021,Teepe2021,Likai2022,Hung-Min2022,Tianrui2022,ren2022,gao2022gait}. According to the related works, the TinySiamese was trained on 60\% of the people in the CASIA-B dataset and tested on the remaining part. 
Table \ref{Tab:casia180Classif} shows the effectiveness of the proposed network for gait classification on three views using GEI images. Table \ref{Tab:casia180Classif2} illustrates similar results compared to various research studies. It is noted that our network was a competitor to the other networks by achieving a high performance.
\begin{table}[ht]
\caption{Classification results of TinySiamese on CASIA-B dataset.}\label{Tab:casia180Classif}
\centering\begin{tabular}{lllll}
\hline
\textbf{Probe Angle} & \multirow{2}{*}{\textbf{Accuracy}} & \multirow{2}{*}{\textbf{Precision}} & \multirow{2}{*}{\textbf{Recall}} & \multirow{2}{*}{\textbf{F1}}\\
\textbf{NM \#5-6} & & & & \\
\hline
 0° & 93.00\% & 94.90\% & 94.90\% & 91.81\%\\
\hline
90° & 100.00\% & 100.00\% & 100.00\% & 100.00\%\\
\hline
180° & 98.98\% & 99.32\% & 98.98\% & 98.91\%\\
\hline
\end{tabular}
\end{table}

\begin{table}[ht]
\caption{Classification accuracies on CASIA-B dataset.}\label{Tab:casia180Classif2}
\centering\begin{tabular}{llll}
\hline
\textbf{Probe Angle NM \#5-6} & \textbf{0°} & \textbf{90°} & \textbf{180°}\\
 \hline
 3DLocal \cite{Huang2021} & 96.0\% & 94.2\% & 95.2\%\\
 \hline
 CSTL \cite{Huang2021(1)} & \textbf{97.8\%} & 95.2\% & 96.7\%\\
 \hline
 GLFE+GLConv \cite{Lin2021} & 96.0\% & 95.4\% & 94.0\%\\
 \hline
 GaitGraph \cite{Teepe2021} & 85.3\% & 86.5\% & 81.9\%\\
 \hline
 GaitSet \cite{Chao2022}& 91.1\% & 94.5\% & 88.0\%\\
 \hline
 STC-Att \cite{Likai2022} & 97.0\% & 97.3\% & 96.0\%\\
 \hline
 GaitTAKE \cite{Hung-Min2022} & 96.7\% & 96.3\% & 96.4\%\\
 \hline
 LagrangeGait \cite{Tianrui2022} & 95.2\% & 94.6\% & 91.5\%\\
 \hline
 RCC-PFL \cite{ren2022} & 87.4\% & 82.6\% & 80.3\%\\
 \hline
 Gait-D \cite{gao2022gait} & 87.7\% & 92.4\% & 87.3\%\\
 \hline
 DenseNet + TinySiamese & 93\% & \textbf{100\%} & \textbf{98.98\%}\\
\hline
\end{tabular}
\end{table}
\section{Discussion}
The proposed TinySiamese architecture obtained not only good results but it also presented a 
short execution time for training and short matching time. These properties allowed for a quick execution, ensuring its use in online verification with the ability to update the network upon new additions without complexity.\\
TinySiamese introduced two new additions compared to standard Siamese networks: the use of a pre-trained model as a feature extractor and a concatenated distance vector in the distance layer.\\
Table 
\ref{Tab:MatchTime} and \ref{Tab:MatchTrain} confirm the effectiveness of this network.
Saving time and memory space are important tasks for quick verification and for people who do not have powerful machines. In fact, the time saved was equal to 612.28 ms for training and 0.086 ms for matching. Despite all these economies, TinySiamese presented encouraging results for verification with an accuracy equivalent to 90.13\% on the fingerprint set of the B2F database, 85.87\% on the face set of the B2F database and the THDD-part1 and 98.39\% on the CASIA-B dataset. Additionally, this network achieved competitive results with related works as shown in table \ref{Tab:casia180Classif2} and \ref{Tab:compfinger}. Thus, the proposed TinySiamese network did not require expensive learning. It can be executed under low-power machines and with fast prediction in terms of execution time.\\
Tables \ref{Tab:FPR} and \ref{Tab:FNR} demonstrate the importance of adding the Hadamard product into the distance layer of the TinySiamese and the Siamese networks. The information added by concatenation with the Euclidean distance of the two extracted feature vectors gave the network more opportunities to avoid mistakes. 
\begin{table}[ht]
\caption{False Positive Rate (FPR).}\label{Tab:FPR}
\centering\begin{tabular}{llll}
\hline
\multirow{2}{*}{Dataset} & \multirow{2}{*}{Method} & \multirow{2}{*}{Euclidean loss} & Euclidean loss\\
& & & + Hadamard Product\\
\hline
B2F & Siamese & 0.073 & 0.060\\
(fingerprint) & TinySiamese & 0.039 &  0.024\\
\hline
B2F (face) & Siamese & 0.033 & 0.052\\
+ THOBRL & TinySiamese & 0.019 &  0.014\\
\hline
\multirow{2}{*}{CASIA-B} & Siamese & 0.0 & 0.076\\
 & TinySiamese & 0.0 &  0.0\\
\hline
\end{tabular}
\end{table}
\begin{table}[ht]
\caption{False Negative Rate (FNR).}
\label{Tab:FNR}
\centering\begin{tabular}{llll}
\hline
\multirow{2}{*}{Dataset} & \multirow{2}{*}{Method} & \multirow{2}{*}{Euclidean loss} & Euclidean loss\\
& & & + Hadamard Product\\
\hline
B2F & Siamese & 0.146 & 0.155\\
(fingerprint) & TinySiamese & 0.106 &  0.171\\
\hline
\hline
B2F (face) & Siamese & 0.405 & 0.307\\
+ THOBRL & TinySiamese & 0.379 &  0.268\\
\hline
\multirow{2}{*}{CASIA-B} & Siamese & 0.607 & 0.360\\
 & TinySiamese & 0.082 &  0.032\\
\hline
\end{tabular}
\end{table}
With all these advantages, it is important to point out two important things. First, the standard Siamese network is able to achieve better results using a bigger number of epochs. In fact, increasing the number of epochs to 1000 gives more chances to the network to learn better. However, this will take more training time and can further strain the computer. Second, the use of convolutional 1D layer instead of the fully-connected layer can reduce the number of parameter in the network more and more. This will be the subject of one of our coming research papers.
\section{Conclusion}
This paper presents a different strategy for performing one-shot verification using the TinySiamese neural network with a 
short matching and training times. The performance of the proposed network was better than the existing state-of-the-art networks developed for biometric verification and classification. The TinySiamese network outperformed the networks available for verification by the lowest complexity and came close to the best results obtained by the previous authors. It presented a saved matching time equal to 76.78\%, a saved training time equivalent to 93.14\% for 10 epochs and a verification accuracy equal to 90.13\% on the fingerprint set of the B2F database, 85.87\% on the face set of the B2F database and the THDD-part1 dataset and 98.39\% on the CASIA-B dataset and a classification accuracy equivalent to 99\% on the FVC2004, 93.00\% for 0° view, 100\% for 90° view and 98.98\% for 180° view on the CASIA-B dataset.
\section*{Acknowledgment}
The research leading to these results has received funding from the Tunisian Ministry of Higher Education and Scientific Research under the grant agreement number LR11ES48.


\bibliographystyle{elsarticle-num} 
\bibliography{elsarticle-template-num}

\pagebreak
\noindent \textbf{Islem Jarraya} was born in Sfax, Tunisia, in 1986. She received the Four-year University Degree in computer science and multimedia in 2009 and the master degree in Computer Science and Multimedia in 2014 at the Multimedia, InfoRmation systems and Advanced Computing Laboratory (MIRACL-Lab) from the higher institute of computer science and multimedia of Sfax, Tunisia. She has been pursuing the PhD. degree on Computing System Engineering with the Research Group on Intelligent Machines (ReGIM-Lab) at ENIS since 2015. She focuses her research on applying intelligent methods (Deep neural networks, feature extraction, and recognition algorithms) to Deep Machine Learning, Computer Vision, Face Image Processing, intelligent pattern recognition, and analysis of large scale complex systems. She has been an active IEEE member since 2015.
\subsection*{  }
\noindent \textbf{Tarek M. Hamdani} (IEEE Member’01, Senior Member'12) was born in Tunis, Tunisia, in 1979. He received his M.Sc. degree in 2003, the Ph.D. degree in 2011, in Computer Science Engineering from the National Engineering School of Sfax, Tunisia, and the HDR degree in Computer Science Engineering from the National Engineering School of Sfax, Tunisia, in 2019. He is currently an Associate Professor in Computer Science at the University of Monastir, Monastir. He focuses his research on applying intelligent methods (neural networks, fuzzy logic, and evolutionary algorithms) to Machine Learning, Computer Vision, Natural Language Processing, intelligent pattern recognition, and analysis of large scale complex systems. He is a Reviewer of the Pattern Recognition Letters, Neurocomputing, Neural processing Letter, IEEE TNNLS, and the Soft Computing journal. He is an IEEE Senior Member of CIS and SMC societies.
\subsection*{  }
\noindent \textbf{Habib Chabchoub} is a full professor of Management Science and director of the MBA program at the College of Business, Abu Dhabi campus, in Al Ain University (UAE). He has initiated and led different Master programs in Management and a PhD program in Quantitative Methods at University of Sfax (Tunisia). He has supervised more than 20 PhD theses , co-worn several international academic and research projects (TEMPUS "Aqi-Umed", CMCU, bilateral projects, and others) and been involved in several international conferences (MOPGP, META, ICALT, LOGISTIQUA, etc.). He co-authored more than 100 refereed publications, several of which are in high impact forums. He received a BSc in Mathematics and a MSc in Management Science from University of Tunis (Tunisia) and a PhD in Operations and Decision Systems from Laval University (Canada). His research interest encompasses supply chain and logistics management, multiple objective programming, multi criteria decision making and meta-heuristics.
\subsection*{  }
\noindent \textbf{Adel M. Alimi} (IEEE Student Member’91, Member’96, Senior Member’00). He graduated in Electrical Engineering in 1990. He obtained a PhD (from Ecole Polytechnique of Montreal, Canada) and then an HDR both in Electrical and Computer Engineering in 1995 and 2000 respectively. He is full Professor in Electrical Engineering at the University of Sfax, ENIS (National Engineering School of Sfax), since 2006. He is founder and director of the research REGIM Lab in intelligent Machines. He was Director of the Tunisia Erasmus+ Office (2018-2020). He also was Director of the National Engineering School of Sfax, University of Sfax, Tunisia (2005-2011). His research interest includes applications of intelligent methods (neural networks, fuzzy logic, evolutionary algorithms) to pattern recognition, robotic systems, vision systems, and industrial processes. He focuses his research on intelligent pattern recognition, learning, analysis and intelligent control of large scale complex systems. He is associate editor and member of the editorial board of many international scientific journals (e.g. “IEEE Trans. Fuzzy Systems”, “NeuroComputing”, “Neural Processing Letters”, “International Journal of Image and Graphics”, “Neural Computing and Applications”, “International Journal of Robotics and Automation”, “International Journal of Systems Science”, etc.). He was guest editor of several special issues of international journals (e.g. Fuzzy Sets \& Systems, Soft Computing, Journal of Decision Systems, Integrated Computer Aided Engineering, Systems Analysis Modelling and Simulations). He is the Founder and Chair of many IEEE Chapter in Tunisia section, he is IEEE Sfax Subsection Chair (2011), IEEE ENIS Student Branch Counselor (2011), IEEE Systems, Man, and Cybernetics Society Tunisia Chapter Chair (2011), IEEE Computer Society Tunisia Chapter Chair (2011), he is also Expert evaluator for the European Agency for Research. He was the general chairman of the International Conference on Machine Intelligence ACIDCA-ICMI’2005 \& 2000.
\end{document}